\pdfoutput=1

\documentclass[11pt]{article}

\usepackage{latex/acl}

\usepackage{graphicx} 
\usepackage{float} 
\usepackage{stfloats}
\usepackage{subfigure} 

\usepackage{pgfplots}
\pgfplotsset{compat=1.18} 
\usepackage{pgfplotstable}

\usepackage{tabularx}
\usepackage{booktabs}
\usepackage{multirow}
\usepackage{array}
\usepackage{color}
\usepackage{tabularray}
\usepackage{bbding}
\usepackage[normalem]{ulem}

\usepackage{amsfonts,amssymb}
\usepackage{amsmath}

\usepackage{times}
\usepackage{latexsym}

\usepackage[T1]{fontenc}

\usepackage[utf8]{inputenc}

\usepackage{microtype}
\usepackage{bbding}
\usepackage{inconsolata}
\title{S$^2$GSL: Incorporating Segment to Syntactic Enhanced Graph Structure Learning for Aspect-based Sentiment Analysis}

\author{
Bingfeng Chen$^1$, Qihan Ouyang$^1$, Yongqi Luo$^1$, Boyan Xu$^1$\thanks{\quad Corresponding authors} , 
Ruichu Cai$^1$, Zhifeng Hao$^{1,2}$ \\
$^1$School of Computer Science, Guangdong University of Technology \\
$^2$College of Science, Shantou University\\
\texttt{chenbingfeng@gdut.edu.cn}\\
\texttt{\{ouyangqihan0720, lyongqi001, hpakyim, cairuichu\}@gmail.com} \\
\texttt{haozhifeng@stu.edu.cn}
}

\begin{document}
\maketitle
\begin{abstract}
Previous graph-based approaches in Aspect-based Sentiment Analysis(ABSA) have demonstrated impressive performance by utilizing graph neural networks and attention mechanisms to learn structures of static dependency trees and dynamic latent trees.
However, incorporating both semantic and syntactic information simultaneously within complex global structures can introduce irrelevant contexts and syntactic dependencies during the process of graph structure learning, potentially resulting in inaccurate predictions.
In order to address the issues above, we propose S$^2$GSL, incorporating \underline{S}egment to \underline{S}yntactic enhanced \underline{G}raph \underline{S}tructure \underline{L}earning for ABSA. 
Specifically, S$^2$GSL is featured with a segment-aware semantic graph learning and a syntax-based latent graph learning enabling the removal of irrelevant contexts and dependencies, respectively.
We further propose a self-adaptive aggregation network that facilitates the fusion of two graph learning branches, thereby achieving complementarity across diverse structures.
Experimental results on four benchmarks demonstrate the effectiveness of our framework.
\end{abstract}

\section{Introduction}
Aspect-based Sentiment Analysis (ABSA) is a fine-grained sentiment analysis task that aims to recognize the sentiment polarities of multiple aspects within a given sentence.
For example, in the sentence "The falafel was rather overcooked and dried but the chicken was fine," the sentiment polarity of the aspect words "falafel" and "chicken" is recognized as negative and positive, respectively. The ABSA task presents a notable challenge in accurately recognizing the sentiment polarity of specific aspect words, particularly when they are influenced by other aspect words with contrasting polarities within the given context.

\setlength{\belowcaptionskip}{-0.7cm}   

\begin{figure}[H] 
\centering 
\includegraphics[width=1.0\columnwidth]{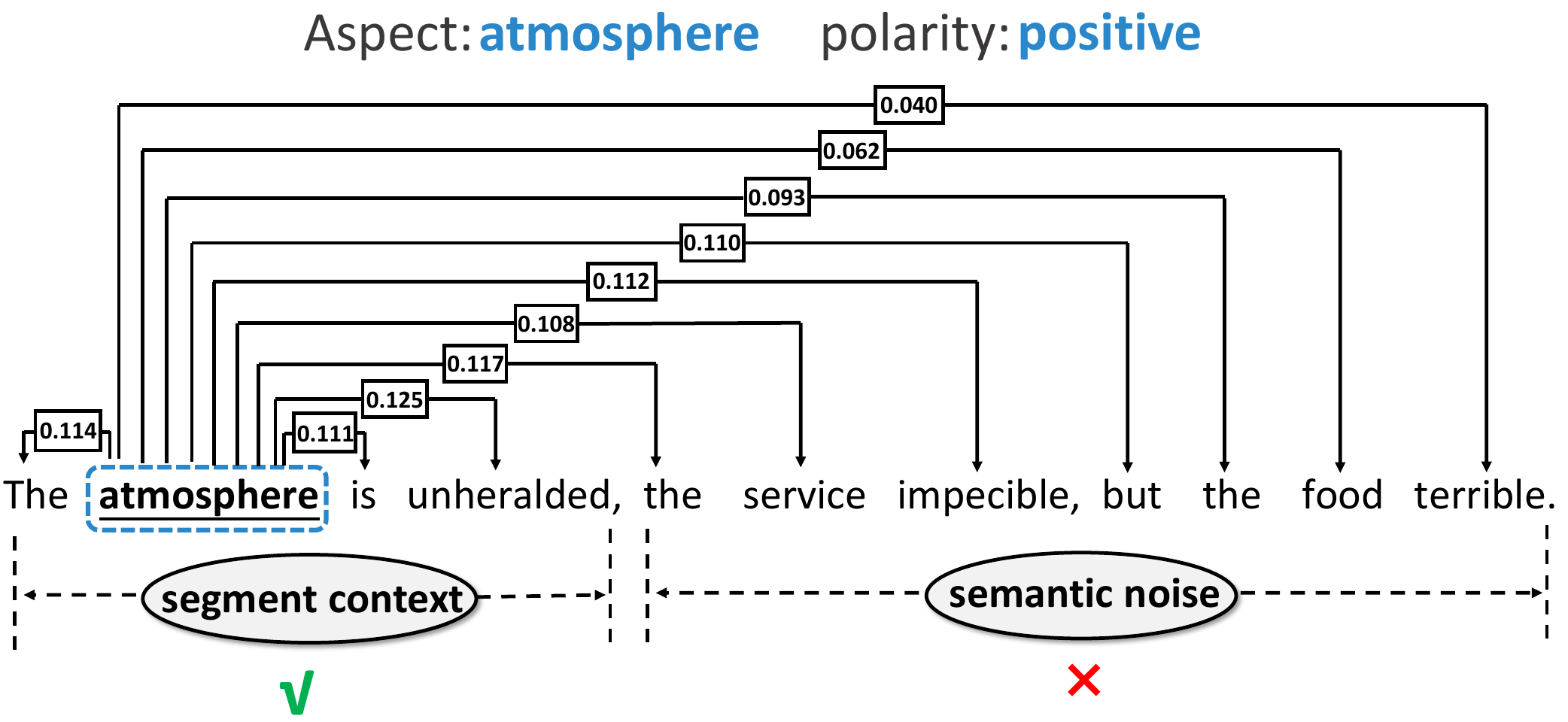} 
\caption{The attention weight of the aspect word "atmosphere" in relation to other words in the sentence."×" refers to noise information for "atmosphere".} 
\label{semantic noise} 
\end{figure}
\setlength{\belowcaptionskip}{-0.2cm}   
\begin{figure}[H] 
\centering 
\includegraphics[width=1.0\columnwidth]{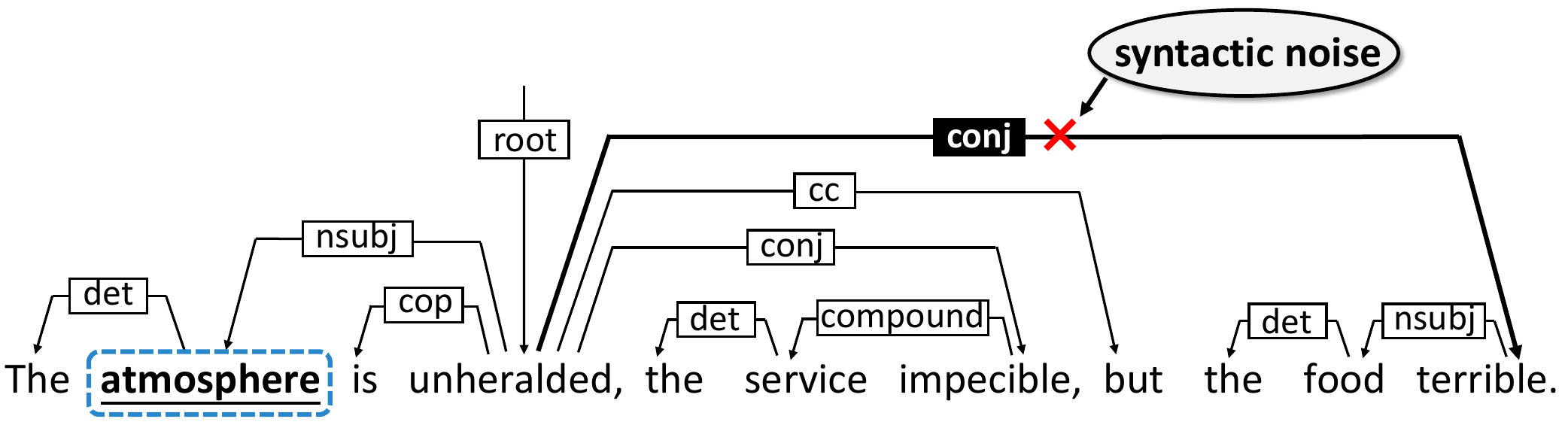} 
\caption{Dependency tree of an example sentence.} 
\label{syntactic noise} 
\end{figure}
Leading graph-based approaches tackle ABSA tasks by learning either prior static structures and learnable dynamic semantic structures via graph neural networks and attention mechanisms. For instance, 
\cite{chen-etal-2020-inducing} proposed combining external syntactic dependency tree and implicit graph to generate aspect-specific representations using GCN;
\cite{li-etal-2021-dual-graph} proposed to construct a SemGCN module and a SynGCN module using an attention mechanism and a syntactic dependency tree.

Although promising results were reported, we observe that existing graph structure learning approaches are still prone to incorrect predictions with hard samples containing multiple aspect words. Existing solutions introduce a heavy structure learning process leading to the following two main limitations: (i) Approaches reliant on attention mechanism are vulnerable to irrelevant context, potentially resulting in misalignment or weak linking. As shown in Figure \ref{semantic noise}, for the aspect word "atmosphere", except for the clause "The atmosphere is unheralded", the other parts are redundant for judging sentiment polarity, i.e., semantic noise. Negatively influenced by irrelevant context, this results in only weak linking to "atmosphere" and the corresponding opinion word "unheralded".
(ii) The global structure of the dependency tree for parsing complex long sentences cannot avoid containing irrelevant dependency information for polarity judgment.
As shown in Figure \ref{syntactic noise}, the "conj" relation connecting "unheralded" and "terrible" is the syntactic noise for the aspect “atmosphere".
Thus, the key to tackling these limitations is efficiently dividing a complex sentence into multiple local clauses in the graph structure learning process.

\begin{figure}[t!] 
\centering 
\includegraphics[width=1.03\columnwidth,height=4.5cm]{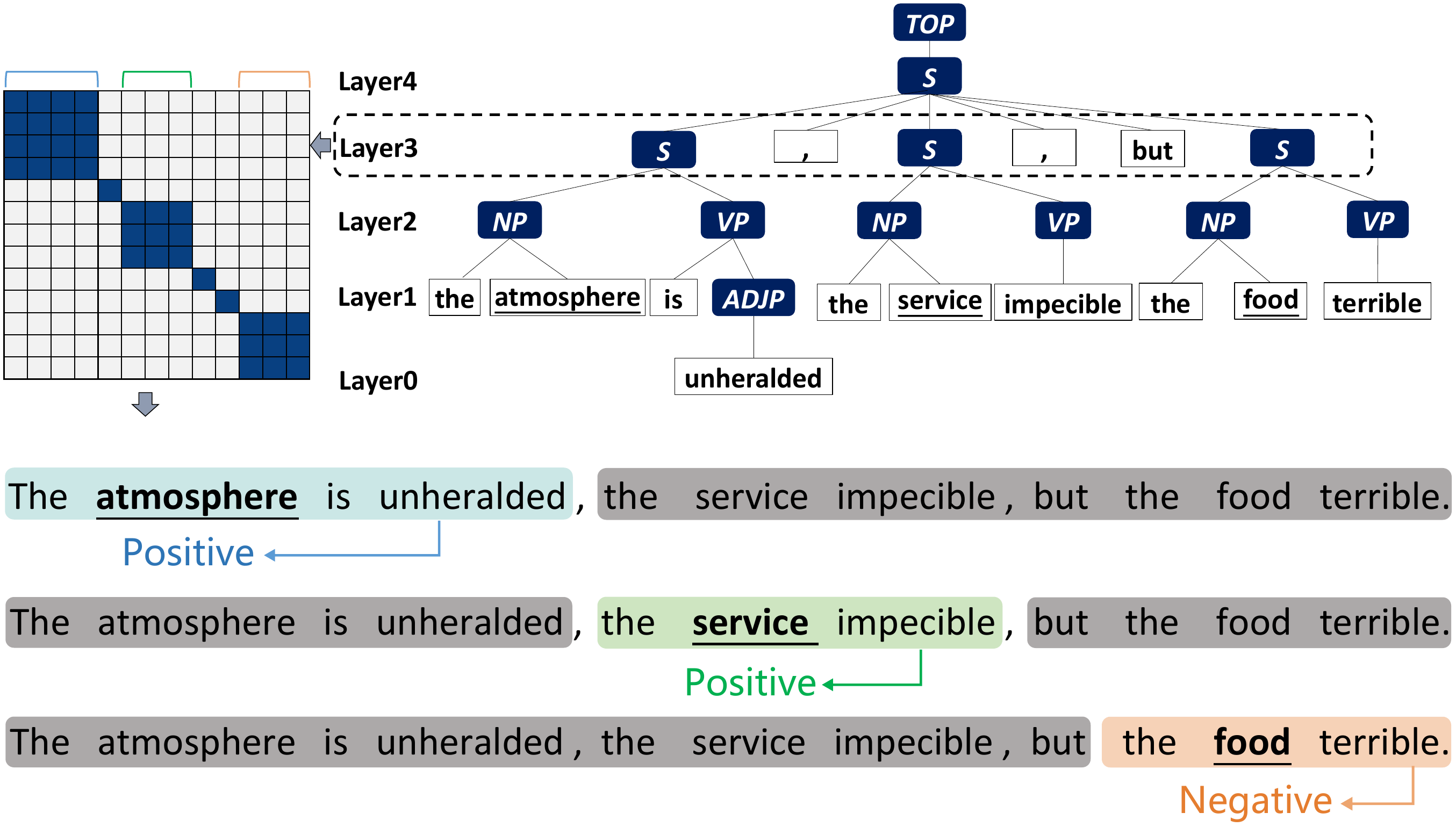} 
\caption{A segment matrix at the third layer of the sample sentence constituent tree divides the sentence into three semantically complete paragraphs.} 
\label{Constituent_tree} 
\end{figure}

In this paper, we propose S$^2$GSL, incorporating \underline{S}egment to \underline{S}yntactic enhanced \underline{G}raph \underline{S}tructure \underline{L}earning for ABSA. To minimize the negative impact of irrelevant structures, S$^2$GSL introduces constituent trees to decompose the complex structure of the input sentences. To illustrate the role of constituent trees, Figure \ref{Constituent_tree} presents a constituent tree and its third-layer segment matrix. This segment matrix can divide a sentence into three semantically complete paragraphs, which help to facilitate the alignment between each aspect and its corresponding opinion and filter out the contextual information unrelated to the respective paragraph.
Specifically, we devise a segment-aware semantic graph(SeSG) branch by using a supervised dynamic local attention on the constituent tree, to learn the local semantic structure of each aspect.
Sharing the same idea with leading graph-based approaches, S$^2$GSL has also been designed with a syntax-based latent graph(SyLG) branch that utilizes syntactic dependency labels to enhance the latent tree construction. The difference from past work is that we introduce an attention-based learning mechanism in SyLG that effectively eliminates irrelevant dependency structures.
Finally, the Self-adaptive Aggregation Network will fuse the SeSG branch and SyLG branch by cross-attention aggregation mechanism, which considers the complementarity across diverse structures. 
Our proposed S$^2$GSL framework makes the following contributions:
\begin{itemize}
\item In contrast to leading approaches in complex graph structure learning for ABSA, our proposed SeSG branch introduces constituent trees to decompose the global structure learning into multiple localized substructure learning processes.

\item In order to reduce dependence on prior structures, our proposed SyLG branch introduces a learnable method to incorporate syntactic dependencies into latent tree construction.
\item Within the two graph learning branch, we propose a Self-adaptive Aggregation Network to facilitate interactions and foster complementary across diverse structures.
\item  We conduct extensive experiments to study the effectiveness of S$^2$GSL. Experiments on four benchmarks demonstrate S$^2$GSL outperforms the baselines. Additionally, the source code and preprocessed datasets used in our work
 are provided on GitHub\footnote{\url{https://github.com/ouy7han/S2GSL}}.

\end{itemize}

\section{Proposed S\texorpdfstring{$^2$}{}GSL}
\begin{figure*}[!htb] 
\centering 
\includegraphics[width=2.0\columnwidth]{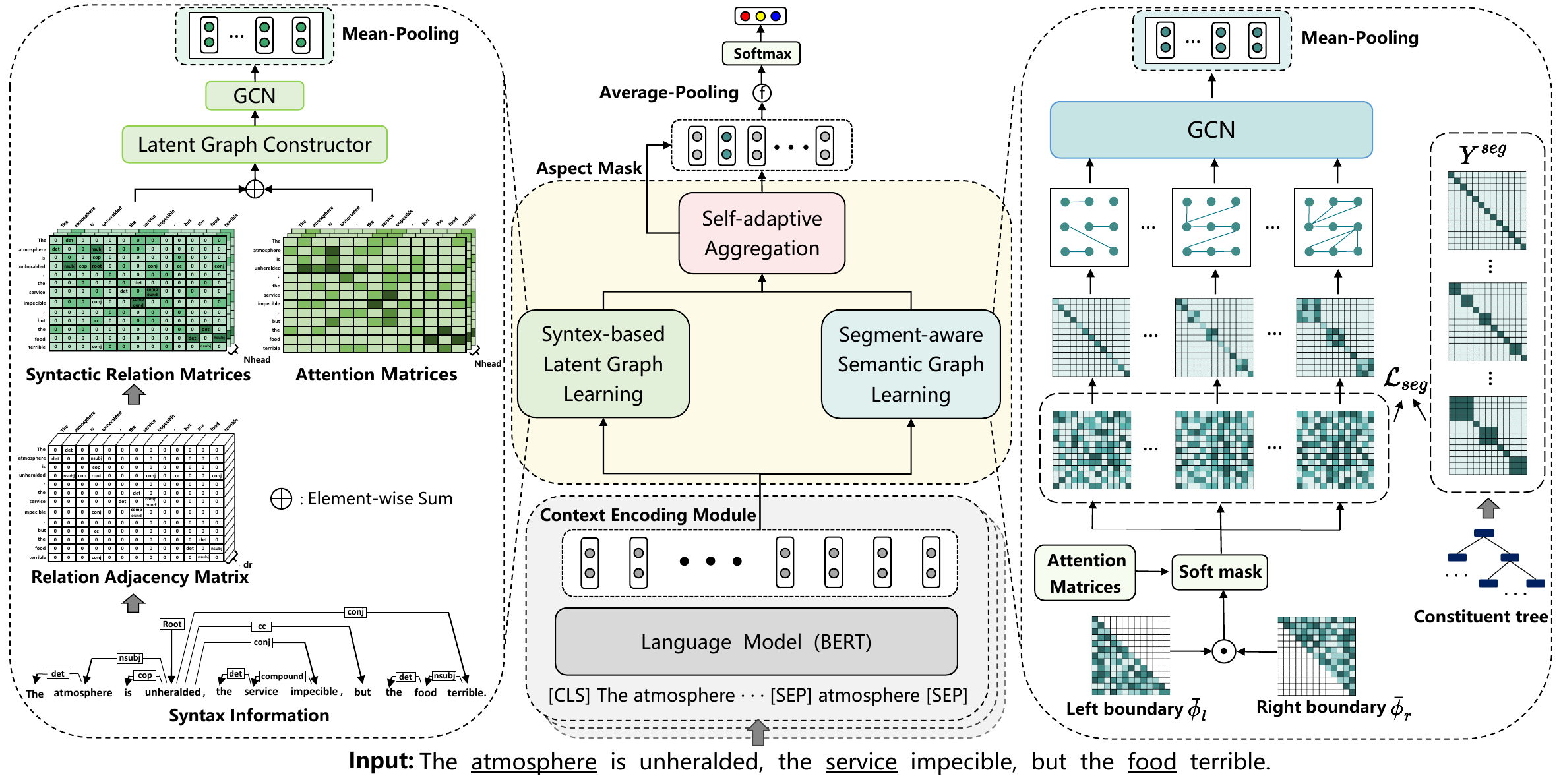} 
\caption{The overall architecture of S$^2$GSL, which is composed primarily of SeSG, SyLG, and Self-adaptive Aggregation Modules.} 
\label{overall_architecture} 
\end{figure*}

The overall architecture of S$^2$GSL is shown in Figure \ref{overall_architecture}
which is mainly composed of four modules: 
Context Encoding Module,
Segment-aware Semantic Graph Learning(SeSG),
Syntax-based Latent Graph Learning(SyLG),
and Self-adaptive Aggregation Module.
Next, components of S$^2$GSL will be introduced separately in the rest of the sections.

\subsection{Context Encoding Module}
Given a sentence of $n$ words 
$s$ = $\{w_{1},w_{2},\dots,$
$w_{\gamma+1}\dots,w_{\gamma+m}\dots,w_{n}\}$
, where the aspect $a$ = $\{w_{\gamma+1},\dots,w_{\gamma+m}\}$,
we use the pre-trained language model BERT \cite{devlin-etal-2019-bert} as sentence encoder to extract contextual representations.
For the BERT encoder, we follow BERT-SPC
\noindent \cite{song2019attentional} to construct a BERT-based sequence
$x$ = ([CLS] $s$ [SEP] $a$ [SEP]), if there are multiple aspects in the sentence, we would construct multiple inputs in the format of $x$.
Then the output representation $H^c$ = \{$h_1^{c},h_2^{c},\dots,h_n^{c}$\} 
$\in$ 
$\mathbb{R}^{n\times d}$ is obtained, where $d$ denotes the dimension of the representation and the $c$ denotes "context".

\subsection{Segment-aware Semantic Graph Learning}
Since aspects are vulnerable to irrelevant context, inspired by recent work\cite{nguyen-etal-2020-differentiable,shang-etal-2021-span}, we use an end-to-end trainable soft masking dynamic local attention mechanism to construct a SeSG branch aiming to align each aspect and its corresponding opinion.

\textbf{Attention Segment Masking Matrix} \ We first generate the word-level attention segments for each sentence by training left and right boundary soft masking matrices
 $\bar{\phi}_{l},\bar{\phi}_{r}\in \mathbb{R}^{n\times n}$, the formulations are calculated as below:
\begin{equation}
\bar{\phi}_{l}=\mathrm{Softmax}\left(\frac{QW_{L}^{Q}(KW_{L}^{K})^{T}}{\sqrt{d}}\odot\hat{M}\right)\label{eq:1} \end{equation}
\begin{equation}
\bar{\phi}_{r}=\mathrm{Softmax}\left(\frac{QW_{R}^{Q}(KW_{R}^{K})^{T}}{\sqrt{d}}\odot\hat{M}^{T}\right)\label{eq:2} \end{equation}
\begin{equation}\left.\hat M_{ij}=\left\{\begin{matrix}1,&i\geq j\\-\infty,&i<j\end{matrix}\right.\right.\label{eq:3} \end{equation}
where $Q$=$K$=$H^c$, $\bigodot$ is the element-wise product,and
$W_{L}^{Q},W_{L}^{K},W_{R}^{Q},W_{R}^{K}\in\mathbb{R}^{d\times d}$
are trainable parameters. 
Notably, a mask matrix $\hat{M}$ is introduced to ensure that the left boundary position $lp$ and the right boundary position $rp$ generated at position $i$ satisfy $0$ $\leq$ $lp$ $\leq$ $i$ $\leq$ $rp$ $\leq$ $N$.

The attention segment masking matrix $M_{s}$ can be obtained by compositing the left and right boundary soft masking matrices $\bar{\phi}_{l}$ and $\bar{\phi}_{r}$ :
\begin{equation}M_{s}=(\bar{\phi}_{l}L_{N})\odot(\bar{\phi}_{r}L_{N}^{T})\end{equation}
where $L_{N}\in\{0,1\}^{n\times n}$ refers to the upper-triangular matrix.

Then we combine the attention segment masking matrix $M_{s}$  with the multi-head attention matrices to enable the model to more focus on the semantically relevant contextual information around each word:
\begin{equation}A^{SeS}=\mathrm{Softmax}\left(\frac{QW^{Q}(KW^{K})^{T}}{\sqrt{d}}\odot M_{s}\right)\end{equation}
where $W^{Q}$, $W^{K}$ are the trainable parameters, $A^{SeS}$ is a multi-head attention matrix with the number of $l$, where $l$ corresponds to the number of layers in the constituent tree.

\textbf{Supervised Constraint} \ In the absence of supervised signal, dynamic local attention may not be able to effectively comprehend the semantically complete segment information around each word, so we further introduce segment-supervised signal to facilitate the learning of dynamic local attention.
Specifically, we use the binary cross-entropy loss to represent the distinction between the attention matrix $A^{SeS}$ and the segment-supervised signal $Y^{seg}$:
\begin{equation}{\mathcal L}_{seg}=BCE(\sigma(A^{SeS}),Y^{seg})\end{equation}
\begin{equation}\left.Y_{ij}^{seg}=\left\{\begin{matrix}1,&S_{ij}=1\\0,&else\end{matrix}\right.\right.\end{equation}
where $\sigma$ represents the sigmoid function, $S_{ij}$ = 1 indicates the $i$-th word and the $j$-th word belong to the same segment, $Y^{seg}$ refers to the segment-supervised signal at each layer of the constituent tree.

To effectively learn the representation of each word, we utilize graph convolutional network \cite{kipf2017semisupervised} to extract the segment-aware semantic features $H^{SeS}$ = \{ $h_1^{SeS}$, $h_2^{SeS}$, \dots, $h_n^{SeS}$\}, which is formulated as below:
\begin{equation}H_{l}^{SeS}=\sigma(A^{SeS}W_{l}H_{l-1}^{SeS}+b_{l})\end{equation}
where $H_{l}^{SeS}$ represents the $l$-th GCN output, $H_{0}^{SeS}=H^{c}$ is the initial input, $\sigma$ denotes a nonlinear activation function, and $W_{l}$ and $b_{l}$ are the trainable parameters.

\subsection{Syntax-based Latent Graph Learning}
Sharing the same idea with past work\cite{tang-etal-2022-affective}, we also adopt syntactic dependency labels to enhance the latent tree construct. The difference from the past work is that we introduce an attention mechanism in the latent tree to effectively eliminate irrelevant dependency structures and construct a SyLG module.

\textbf{Syntactically Enhanced Weight Matrix} \ 
To leverage dependency label information, we first use an off-the-shelf toolkit to obtain dependency information, then we utilize this information to generate a dependency type matrix $R=\left\{r_{i,j}\right\}_{n\times n}$, where $r_{i,j}$ represents the types of dependency between $x_{i}$ and $x_{j}$  \cite{tian-etal-2021-aspect}. Subsequently, we embed each dependency type $r_{i,j}$ into the vector $e_{ij}\in\mathbb{R}^{1\times dr}$, and finally obtain the relational adjacency matrix $M_{R}=\left\{e_{ij}|1\leq i\leq n,1\leq j\leq n\right\}$, where $e_{ij}$ refers to the embedding vector of dependency type between the $i$-th word and the $j$-th word. If $w_{i}$ and $w_{j}$ are not connected, we assign a "0" embedding vector to $e_{ij}$.

In order to induce a syntax-based latent tree, we need to generate a syntactically enhanced weight matrix.
Specifically, we first use multi-head self-attention mechanism to compute a weight matrix $A_{a}$. Then, we transform the relation adjacency matrix $M_{R}$ into a syntactic relation weight matrix $A_{r}$ through a linear transformation, which has the same number of heads as $A_{a}$. Finally, the syntactically enhanced weight matrix $\bar{A}$ can be obtained by summing $A_{a}$ and $A_{r}$:
\begin{equation}A_a^\mathrm{k}=\text{softmax}\left(\frac{QW^Q\times(KW^K)^\mathrm{T}}{\sqrt{d}}\right)\end{equation}
\begin{equation}A_{r}=(W_{R}^{T}M_{R}+b_{R})\end{equation}
\begin{equation}\bar{A}=\mathrm{softmax}\left(A_{r}+A_{a}\right)\end{equation}
where $A_{a}^{k}$ is the attention score matrix of the $k$-th head, $W_{R}\in\mathbb{R}^{d_r\times N_{head}}$ is the weight matrix for the linear transformation.

\textbf{Syntax-based Latent Tree Construction} \ Considering $\bar{A}$ as the initial weight matrix, we follow \cite{zhou-etal-2021-closer} to generate a syntax-based latent tree. We firstly  define a variant of the Laplacian matrix for the syntax-based latent tree:
\begin{equation}\bar{L}_{ij}=\begin{cases}\Phi_{i}+\sum_{i'=1}^n\bar{A}_{i'j}&\text{if} \ i=j\\-\bar{A}_{ij}&\textit{otherwise}\end{cases}\end{equation}
 where $\Phi_{i}=\exp({W}_{r}{h}_{i}^{c}+{b}_{r})$ is the non-normalized score that the $i$-th node is selected as the root node, $\bar{L}$ can be used to simplify the computation of weight sums. Therefore, the marginal probability $A_{ij}^{SyL}$ of the syntax-based latent tree can be computed using $\bar{L}_{ij}$:
\begin{equation}\begin{aligned}{A}^{SyL}_{ij}&=\left(1-\delta_{1,j}\right)\bar{A}_{ij}\left[\bar{L}^{-1}\right]_{jj}\\&-\left(1-\delta_{i,1}\right)\bar{A}_{ij}\left[\bar{L}^{-1}\right]_{ji}\end{aligned}\end{equation}
where $\delta$ is Kronecker delta, $A^{SyL}$ can be regarded as the adjacency matrix of the syntax-based latent tree. 

We employ a root constraint strategy \cite{zhou-etal-2021-closer} to direct the root node towards the aspect word:
\begin{equation}\mathcal{L}_{r}=-\sum_{i=1}^{N}(t_{i}\mathrm{log}\boldsymbol{P}_{i}^{r})+(1-t_{i})\mathrm{log}\left(1-\boldsymbol{P}_{i}^{r}\right)\end{equation}
where $\boldsymbol{P}_{i}^{r}=\Phi_{i}[\bar{L}^{-1}]_{i1}$ denotes the probability of the $i$-th word being the root node, and $t_{i}$ $\in$ \{0, 1\} indicates whether the $i$-th word is an aspect word.

Similar to the SeSG, we utilize graph convolutional network to extract the syntax-based latent Graph features $H^{SyL}$ = \{ $h_1^{SyL}$, $h_2^{SyL}$, \dots, $h_n^{SyL}$\}, which is formulated as below:
\begin{equation}H_{l}^{SyL}=\sigma(A^{SyL}W_{l}H_{l-1}^{SyL}+b_{l})\end{equation}
where $H_{l}^{SyL}$ denotes the $l$-th layer of GCN output, $H_{0}^{SyL}=H^{c}$ is the initial input, $W_{l}$ and $b_{l}$ are the trainable parameters.

\subsection{Self-adaptive Aggregation Module}
Considering the complementarity between SeSG and SyLG, we design a self-adaptive aggregation 
\begin{figure}[H] 
\centering 
\includegraphics[width=1.0\columnwidth]{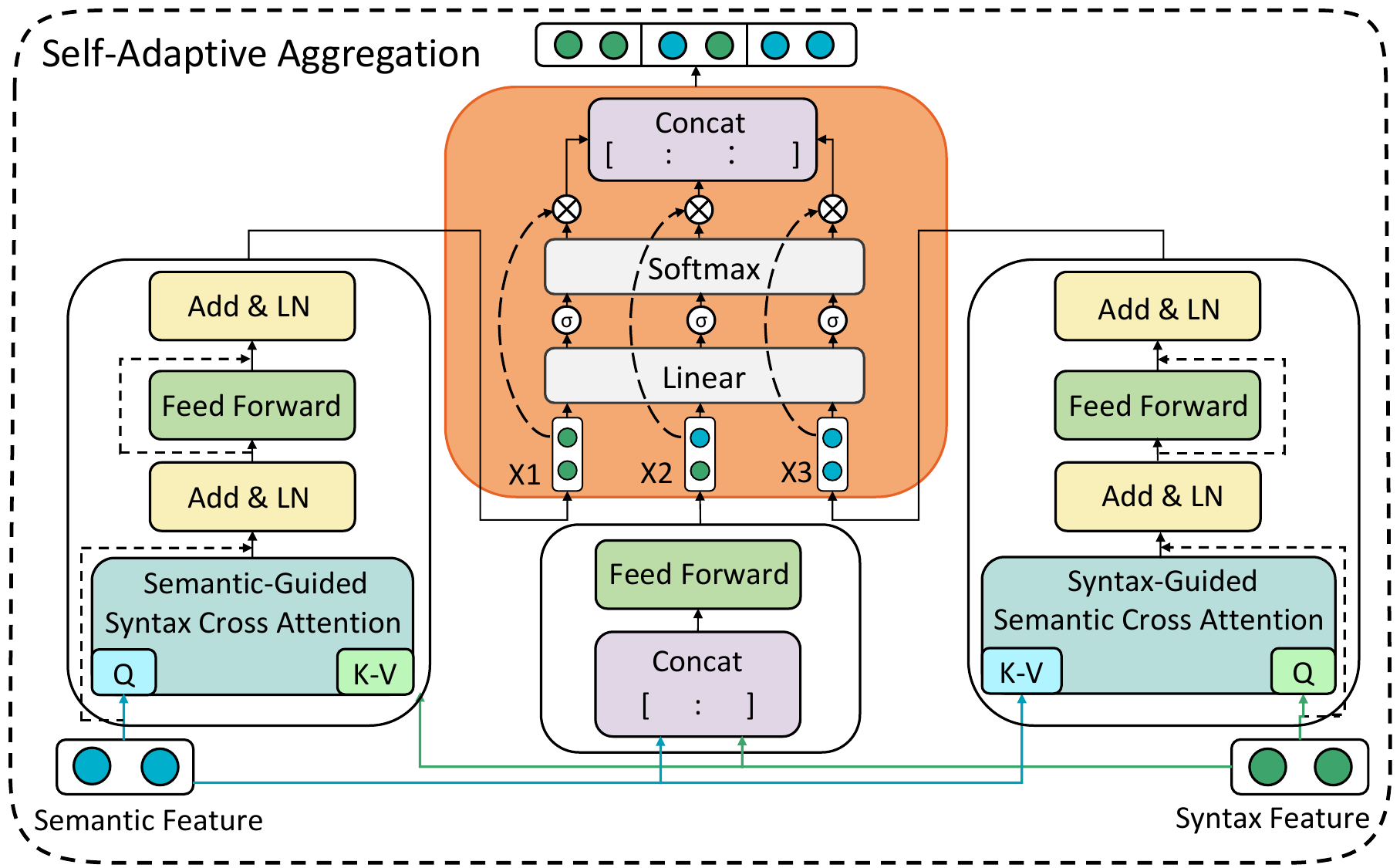} 
\caption{The overall architecture of Self-adaptive Aggregation Module.} 
\label{Self-adaptive Fusion} 
\end{figure}
\noindent module, as shown in Figure \ref{Self-adaptive Fusion}, to realize their interaction. Specifically, this module consists of three streams, with two of them extended from the traditional Transformer \cite{10.5555/3295222.3295349}. These two streams can combine information from SeSG and SyLG and ultimately obtain semantic-guided syntax representations $H^{SemG}$ and syntax-guided semantic representations $H^{SynG}$:
\begin{equation}\begin{aligned}O^{Sem}&=LN(MH(Q=H^{SeS},K=V\\& \quad =H^{SyL})+H^{SeS})\end{aligned}\end{equation}
 \begin{equation}H^{SemG}=LN(FFN(O^{Sem})+O^{Sem})\end{equation}
\begin{equation}\begin{aligned}O^{Syn}&=LN(MH(Q=H^{SyL},K=V\\& \quad =H^{SeS})+H^{SyL})\end{aligned}\end{equation}
\begin{equation}H^{SynG}=LN(FFN(O^{Syn})+O^{Syn})\end{equation}
where $H^{SemG}, H^{SynG} \in R^{n\times d}$, $MH(\cdot)$ denotes multi-head attention, $LN(\cdot)$ refers to layer normalization, and $FFN(\cdot)$ represents the feed-forward neural network.

To avoid bias towards specific module information, we introduce an additional channel to balance the information between SeSG and SyLG. The specific approach is as follows:
\begin{equation}H^{Com}=FFN(Concat([H^{SeS},H^{SyL}]))\end{equation}
where $H^{Com}\in R^{n\times d}$, $Concat(\cdot)$ represents the concatenation function.

Considering the different roles of the various module outputs, we assign different weights to these 
outputs to allow the model to more focus on the important module.
Technically, given the input features $X=[X_{1},X_{2},X_{3}]$. The weight of each module is calculated by the following equation:
\begin{equation}a_i=\mathrm{ReLU}(W^TX_i+b)\end{equation}
\begin{equation}\alpha_{i}=\frac{exp(a_{i})}{\sum_{j=1}^{3}exp(a_{j})}\end{equation}
where $X_{1}=H^{SemG}$, $X_{2}=H^{SynG}$, $X_{3}=H^{Com}$, $W_{l}$ and $b_{l}$ are the trainable parameters.The final output feature $H^{F}$ is generated as follows:
\begin{equation}H^{F}=Concat([\alpha_{1}H^{SemG},\alpha_{2}H^{SynG},\alpha_{3}H^{Com}])\end{equation}
where $H^{F}\in R^{n\times d_{3}}$, with $d_{3}$ = 3$d$.

\subsection{Training}
We use average pooling at the final aspect nodes of $H^{F}$ to obtain the aspect representation $H^{a}$. Then, the sentiment probability distribution $y_{(s, a)}$ is calculated using a linear layer with a softmax function:
\begin{equation}y_{(s,a)}=\mathrm{softmax}(W^{p}{H}^{a}+b^{p})\end{equation}
where ($s$,$a$) represents the sentence-aspect pair. Our training objective is to minimize the following objective function:
\begin{equation}\mathcal{L}(\Theta)=\mathcal{L}_{C}+\lambda_{1}\mathcal{L}_{seg}+\lambda_{2}\mathcal{L}_{r}\end{equation}
where $\Theta$ denotes all trainable parameters of the model, $\lambda_{1}$ and $\lambda_{2}$ are hyper-parameters, and $\mathcal{L}_{C}$ is the standard cross-entropy loss function:
\begin{equation}\mathcal{L}_{\mathrm{C}}=-\sum_{(s,a)\in D}\sum_{c\in C}\log y_{(s,a)}\end{equation}
where $D$ contains all sentence-aspect pairs and $C$ is the collection of different sentiment polarities.

\section{Experiments}

\subsection{Datasets}
We conduct experiments on four public datasets. 
The Restaurant and Laptop reviews are from SemEval 2014 Task 4 \cite{pontiki-etal-2014-semeval}. The Twitter dataset is a collection of tweets \cite{dong-etal-2014-adaptive}. The MAMS dataset is consisted of sentences with multiple aspects\cite{jiang-etal-2019-challenge}.
Each aspect in the sentence is labeled with one of the three sentiment polarities: positive, neutral, and negative. The statistics for the four datasets are shown in Table \ref{table11}.

\begin{table}[H]
\centering
\scalebox{0.8}{\begin{tblr}{
  cells = {c},
  cell{1}{1} = {r=2}{},
  cell{1}{2} = {c=2}{},
  cell{1}{4} = {c=2}{},
  cell{1}{6} = {c=2}{},
  hline{1,7} = {-}{0.08em},
  hline{2} = {2-7}{},
  hline{3-6} = {-}{},
}
\textbf{Dataset} & \textbf{\#Positive} &      & \textbf{\#Negative} &      & \textbf{\#Neutral} &      \\
                 & Train               & Test & Train               & Test & Train              & Test \\
Laptop           & 976                 & 337  & 851                 & 128  & 455                & 167  \\
Restaurant       & 2164                & 727  & 807                 & 196  & 637                & 196  \\
Twitter          & 1507                & 172  & 1528                & 169  & 3016               & 336  \\
MAMS             & 3380                & 400  & 2764                & 329  & 5042               & 607  
\end{tblr}}
\caption{Statistics for the four experimental datasets.}
\label{table11}
\end{table}

\subsection{Implementation Details}
The Stanford parser\footnote{\url{https://stanfordnlp.github.io/CoreNLP/}} \cite{manning-etal-2014-stanford} is used to obtain syntactic dependencies.
Specifically, we use CRF constituency parser \cite{ijcai2020p560} to obtain the constituent tree.
We use the bert-base-uncase\footnote{\url{https://github.com/huggingface/transformers}}  model as our context encoder.
The model training is conducted using the Adam optimizer 
with a learning rate of $2\times10^{-5}$ and L2 regularization of $10^{-5}$.
The GCN layers of SeSG and SyLG are set to 3.
Our model is trained in 20 epochs with a batch size of 16.
The hyper-parameters $\lambda_{1}$ and $\lambda_{2}$  for the four datasets are (0.1,0.5), (0.1,0.45), (0.35,0.3) and (0.4,0.75).
All experiments are conducted on an NVIDIA 3090 GPU. The model with the highest accuracy or F1 score among all evaluation results is selected as the final model.

\begin{table*}[h!t]
\centering
\scalebox{0.72}{\begin{tblr}{
  cells = {c},
  cell{1}{1} = {r=2}{},
  cell{1}{2} = {c=2}{},
  cell{1}{4} = {c=2}{},
  cell{1}{6} = {c=2}{},
  cell{1}{8} = {c=2}{},
  cell{3}{8} = {c=2}{},
  cell{4}{8} = {c=2}{},
  cell{6}{8} = {c=2}{},
  cell{7}{8} = {c=2}{},
  cell{9}{8} = {c=2}{},
  cell{10}{8} = {c=2}{},
  cell{12}{8} = {c=2}{},
  cell{13}{8} = {c=2}{},
  hline{1,15} = {-}{0.08em},
  hline{2} = {2-9}{},
  hline{3,14} = {-}{},
}
\textbf{Model } & \textbf{Laptop}   &                   & \textbf{Restaurant} &                   & \textbf{Twitter}  &                   & \textbf{MAMS}     &                   \\
                & \textbf{Accuracy} & \textbf{Macro-F1} & \textbf{Accuracy}   & \textbf{Macro-F1} & \textbf{Accuracy} & \textbf{Macro-F1} & \textbf{Accuracy} & \textbf{Macro-F1} \\
RAM \cite{chen-etal-2017-recurrent}             & 74.49             & 71.35             & 80.23               & 70.80             & 69.36             & 67.30             & -                 &                   \\
MGAN \cite{fan-etal-2018-multi}            & 75.39             & 72.47             & 81.25               & 71.94             & 72.54             & 70.81             & -                 &                   \\
R-GAT \cite{wang-etal-2020-relational}           & 78.21             & 74.07             & 86.60               & 81.35             & 76.15             & 74.88             & 84.52             & 83.74             \\
KumaGCN \cite{chen-etal-2020-inducing}         & \uline{81.98}     & \uline{78.81}     & 86.43               & 80.30             & 77.89             & 77.03             & -                 &                   \\
ACLT \cite{zhou-etal-2021-closer}            & 79.68             & 75.83             & 85.71               & 78.44             & 75.48             & 74.51             & -                 &                   \\
T-GCN \cite{tian-etal-2021-aspect}           & 80.88             & 77.03             & 86.16               & 79.95             & 76.45             & 75.25             & 83.38             & 82.77             \\
DualGCN \cite{li-etal-2021-dual-graph}         & 81.80             & 78.10             & 87.13               & 81.16             & 77.40             & 76.02             & -                 &                   \\
SSEGCN \cite{zhang-etal-2022-ssegcn}          & 81.01             & 77.96             & 87.31               & 81.09             & 77.40             & 76.02             & -                 &                   \\
dotGCN \cite{chen-etal-2022-discrete}          & 81.03             & 78.10             & 86.16               & 80.49             & \uline{78.11}     & 77.00             & \uline{84.95}             & \uline{84.44}             \\
MGFN  \cite{tang-etal-2022-affective}           & 81.83             & 78.26             & \uline{87.31}       & \uline{82.37}     & \textbf{78.29}    & \textbf{77.27}    & -                 &                   \\
TF-BERT(dec) \cite{zhang-etal-2023-span}    & 81.49             & 78.30             & 86.95               & 81.43             & 77.84             & 76.23             & -                 &                   \\
\textbf{S$^2$GSL(Ours)}          & \textbf{82.46}    & \textbf{79.07}    & \textbf{87.31}      & \textbf{82.84}    & 77.84             & \uline{77.11}     & \textbf{85.17}    & \textbf{84.74}    
\end{tblr}}
\caption{The main experimental results on four public datasets. The best are in bold, and second-best are underlined.}
\label{table2}
\end{table*}

\begin{table*}[h!t]
\centering
\scalebox{0.74}{\begin{tblr}{
  row{odd} = {c},
  row{4} = {c},
  row{6} = {c},
  cell{1}{1} = {r=2}{},
  cell{1}{2} = {c=2}{},
  cell{1}{4} = {c=2}{},
  cell{1}{6} = {c=2}{},
  cell{1}{8} = {c=2}{},
  cell{2}{8} = {c},
  cell{2}{9} = {c},
  hline{1,3-4,7} = {-}{},
  hline{2} = {2-9}{},
}
\textbf{Model}                 & \textbf{Laptop}   &                   & \textbf{Restaurant} &                   & \textbf{Twitter } &                   & \textbf{MAMS}     &                   \\
                               & \textbf{Accuracy} & \textbf{Macro-F1} & \textbf{Accuracy}   & \textbf{Macro-F1} & \textbf{Accuracy} & \textbf{Macro-F1} & \textbf{Accuracy} & \textbf{Macro-F1} \\
\textbf{\textbf{S$^2$GSL(Ours)}}  & \textbf{82.46}    & \textbf{79.07}    & \textbf{87.31}      & \textbf{82.84}    & \textbf{77.84}    & \textbf{77.11}    & \textbf{85.17}    & \textbf{84.74}    \\
w/o SyLG                      & 80.41             & 77.39             & 86.50               & 80.22             & 75.92             & 74.87             & 83.71             & 83.23             \\
w/o SeSG                      & 79.46             & 76.33             & 86.10               & 79.66             & 76.51             & 75.23             & 83.42             & 82.76             \\
w/o Self-Adaptive Aggregation & 80.88             & 77.07             & 86.32               & 80.08             & 76.07             & 75.54             & 83.60             & 83.05             
\end{tblr}}
\caption{Experimental results of ablation study.}
\label{table3}
\end{table*}
\subsection{Baselines}
We compare our S$^2$GSL with some mainstream and lasted models in ABSA, including 
\textbf{Attention-based methods}: RAM
\cite{chen-etal-2017-recurrent}, MGAN \cite{fan-etal-2018-multi}.
\textbf{Syntactic-based methods}: R-GAT\cite{wang-etal-2020-relational},T-GCN\cite{tian-etal-2021-aspect}.
\textbf{Latent-graph methods}: ACLT \cite{zhou-etal-2021-closer}, dotGCN \cite{chen-etal-2022-discrete}.
\textbf{Multi-graph combined methods}:
KumaGCN \cite{chen-etal-2020-inducing}, DualGCN \cite{li-etal-2021-dual-graph}, SSEGCN\cite{zhang-etal-2022-ssegcn}
MGFN \cite{tang-etal-2022-affective}.
\textbf{Other method}:
TF-BERT \cite{zhang-etal-2023-span}.

\subsection{Overall Performance}
All baseline results on four datasets are shown in Table \ref{table2}, we can find that our S$^2$GSL outperforms all baselines on Laptop, Restaurant, and MAMS datasets. We got the second best on the Twitter dataset, reaching comparable with MGFN. We guess it is because the sentence structure of Twitter dataset is more complex than the other three datasets, basically samples containing only single aspect words, which cannot reflect the advantages of S$^2$GSL. In contrast, the other three datasets,
laptop, restaurant, and MAMS contain samples of multiple aspect words. The effect enhancement in  Laptop, Restaurant, and MAMS effectively supports that S$^2$GSL gets better sentiment recognition in the case of having multiple aspect words. Additionally, we conduct a parameter comparison between S$^2$GSL and the GCN-based baseline methods. Notably, the number of parameters in S$^2$GSL is comparable (detailed results can be found in \ref{A.4}). These results demonstrate that S$^2$GSL exhibits superior performance while incurring the same computational overhead.


\begin{figure}[H] 
\centering 
\includegraphics[width=1.03\columnwidth]{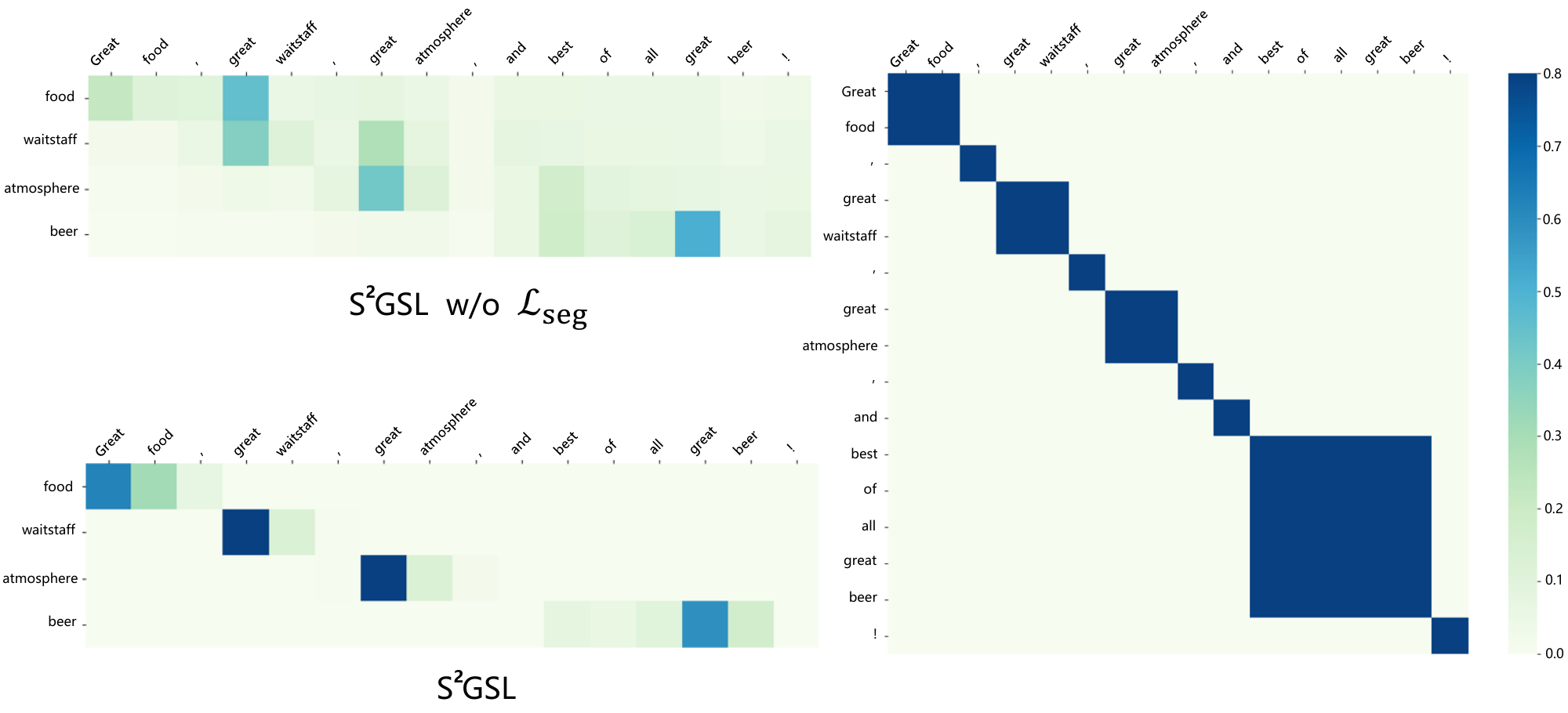} 
\caption{Visualization of attention weights for all aspects (left) and segment-supervised signal (right).} 
\label{supervised dynamic local attention} 
\end{figure} 
\subsection{Ablation Study}
We conduct ablation experiments to further investigate the effects of different modules, 
shown in Table \ref{table3}.
Excluding the syntax-based latent graph learning (w/o SyLG) results in a decrease in the model's performance, which demonstrates the importance of the ability to adaptively capture syntactic relationships between words.
We remove the segment-aware semantic graph learning module (w/o SeSG). Compared to the Twitter dataset, the performance of the Restaurant, Laptop and MAMS datasets 
decreases significantly, which is due to the fact that the sentence structures on these three datasets are more formal and each sentence can be constructed with multiple subordinate clauses.
w/o Self-Adaptive Aggregation refers to the SyLG and SeSG modules cannot interact with each other, leading to a drop in performance on all four datasets. The study reveals that both SyLG and SeSG branches are crucial for handling complex sentences in all datasets, as removing either component leads to a noticeable drop in performance. However, compared to the other three datasets, the performance of the SeSG module on the Twitter dataset is not particularly significant, since each sentence on Twitter contains only one aspect word. This difference underscores the adaptability and effectiveness of S$^2$GSL in varying complexity levels across datasets.

\section{Discuss and Analysis}
\subsection{Effect of Dynamic Local Attention}
To demonstrate the effectiveness of supervised dynamic local attention, we visualize the attention weight of all aspects in a sentence, as shown in Figure \ref{supervised dynamic local attention}.
We can find that in the model without the constraints of segment-supervised signal (S$^2$GSL w/o $\mathcal{L}_{seg}$), the aspect "food" incorrectly assigns a higher attention weight to the opinion "great" of "waitstaff". In contrast, attention weights of each 
\begin{figure}[H]
\centering  
\subfigure[Dependency tree of an example sentence. Respective aspects and opinions are connected through the "nsubj" label.]{
\label{dependence tree}
\includegraphics[width=1.0\columnwidth]{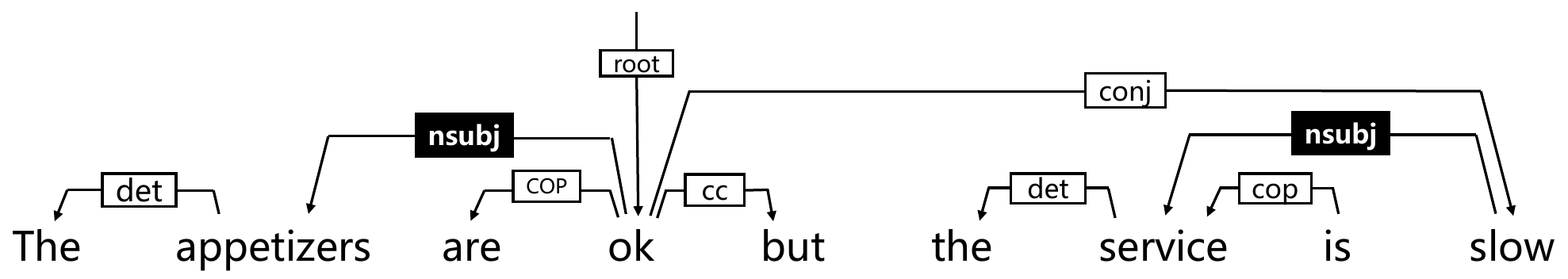}}
\subfigure[The weight change of each aspect word assigned to the corresponding opinion.]{
\label{syntax-based latent tree}
\includegraphics[width=1.0\columnwidth]{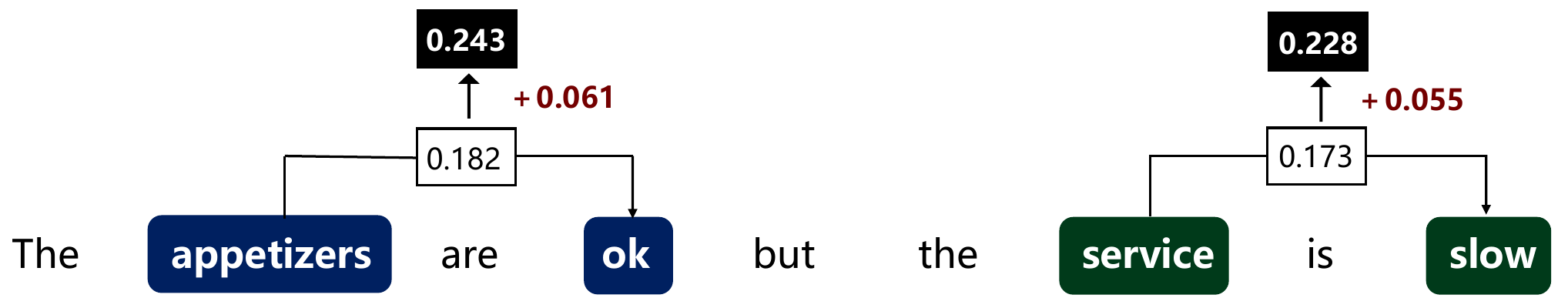}}
\caption{Effect of syntactic dependency label.}
\label{latent tree}
\end{figure}

\pgfplotstableread[col sep=comma]{
	Dataset, concat, sum, gate, ours
	Laptop,80.09,79.93,80.88,82.46
	Restaurant,86.59,86.14,86.77,87.31
	Twitter,76.21,76.46,77.10,77.84
	}\mytable
\definecolor{blueaccent}{RGB}{0,150,214}
\definecolor{greenaccent}{RGB}{0,139,43}
\definecolor{purpleaccent}{RGB}{130,41,128}
\definecolor{orangeaccent}{RGB}{240,83,50}
\begin{figure}[H] 
\centering 
\begin{minipage}[t]{0.30\columnwidth}
\centering
 \begin{tikzpicture}[scale=.49]
  \begin{axis}[
    width=7cm,
    ybar,
    bar width=7pt,
    ymin=74,
    enlarge x limits={abs=27pt},
    legend style={draw=none,at={(0.5,-0.15)},
        anchor=north,legend columns=-1},
    ylabel={\textbf{Accuracy.($\%$)}},
    symbolic x coords={Laptop,Restaurant,Twitter},
    ytick={74,76,78,80,82,84,86,88},
    xtick=data,
    cycle list={blueaccent,greenaccent,purpleaccent,orangeaccent}
  ]
    \pgfplotsinvokeforeach{concat, sum, gate, ours}{
      \addplot+[draw=none,fill,] table[x=Dataset,y=#1]{\mytable};
      \addlegendentry{#1}}
\end{axis}
\end{tikzpicture}
\end{minipage}
\quad
\pgfplotstableread[col sep=comma]{
	Dataset, concat, sum, gate, ours
	Laptop,76.60,75.30,77.34,79.07
	Restaurant,80.21,79.66,81.38,82.84
	Twitter,74.64,75.16,76.03,77.11
	}\mytable
\begin{minipage}[t]{0.63\columnwidth}
\centering
\begin{tikzpicture}[scale=.49]
\begin{axis}[
    width=7cm,
    ybar,
    bar width=7pt,
    ymin=72,
    enlarge x limits={abs=27pt},
    legend style={draw=none,at={(0.5,-0.15)},
        anchor=north,legend columns=-1},
    ylabel={\textbf{F1.($\%$)}},
    symbolic x coords={Laptop,Restaurant,Twitter},
    ytick={70,72,74,76,78,80,82,84},
    xtick=data,
    cycle list={blueaccent,greenaccent,purpleaccent,orangeaccent}
  ]
    \pgfplotsinvokeforeach{concat, sum, gate, ours}{
      \addplot+[draw=none,fill,] table[x=Dataset,y=#1]{\mytable};
      \addlegendentry{#1}}
\end{axis}
\end{tikzpicture}
\end{minipage}
\caption{Effects of different fusion strategies.} 
\label{different fusion strategies} 
\end{figure}
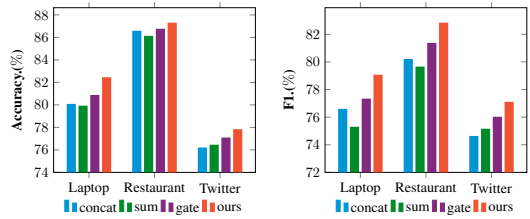
\noindent aspect word constrained by the supervised signals are concentrated in its semantically coherent segment and each aspect assigns a higher attention weight to its corresponding opinion, which helps avoid the influence of noisy information.

\subsection{Effect of Syntactic Dependency Label}
To investigate the validity of syntactic dependency label information, we analyzed the weight change of each aspect word assigned to the corresponding opinion word in the latent tree. Figure \ref{syntax-based latent tree} shows the weight changes of the aspect words "appetizers" and "service" assigned to the corresponding opinion "ok" and "slow".  As can be seen from the figure, their respective weights have increased by 0.061 and 0.055. This is because in most cases, each aspect is connected to its corresponding opinion through the same syntactic dependency label, such as the "nsubj" label in Figure \ref{dependence tree}. The above analysis indicates that dependency labels can better capture the relationship between aspects and their corresponding opinions.

\subsection{Effects of different fusion strategies}
To validate the effectiveness of our proposed self-adaptive aggregation module, we compare it with several typical information fusion strategies: "concat", "sum" and "gate".
As shown in Figure \ref{different fusion strategies}, we can find that "gate" outperforms better than "concat" and "sum" on all datasets. Furthermore, "con-
\begin{table}[H]
\centering
\scalebox{0.8}{\begin{tblr}{
  cells = {c},
  cell{1}{1} = {r=2}{},
  hline{1,8} = {-}{0.08em},
  hline{2} = {2-3}{},
  hline{3,7} = {-}{},
}
\textbf{Model}       & \textbf{Laptop14} & \textbf{Restaurant14} \\
                     & \textbf{F1}       & \textbf{F1}           \\
5-shot ICL \cite{Han2023IsIE}           & 76.76             & 81.85                 \\
Prompt-setting 1     & 75.62             & 79.48                 \\
Prompt-setting 2     & 77.02             & 81.38                 \\
Prompt-setting 3     & 77.91             & 82.14                 \\
\textbf{S$^2$GSL(Ours)} & \textbf{79.09}    & \textbf{82.84}        
\end{tblr}}
\caption{The results obtained using ChatGPT and S$^2$GSL.}
\label{table44}
\end{table}
\noindent cat" performs better than "sum" on the laptop and restaurant, while "sum" performs better on Twitter. These results provide evidence that direct fusion strategies (e.g., concat and sum) are sub-optimal. In contrast, our proposed fusion module achieved the best performance on all datasets, which proves that our fusion module adaptively fuses the respective information in a multi-stream manner, which can fully utilize the complementarities between each stream.

\subsection{Experiments With ChatGPT}
ChatGPT \cite{openai2023}, powered by GPT-3.5 and GPT-4, can achieve significant zero-shot and few-shot in-context learning (ICL) \cite{Brown2020LanguageMA} performance on unseen tasks, even without any parameter updates.

In this section, we investigate the performance of ChatGPT on ABSA tasks and its ability for fine-grained understanding of segmental context. We experimented with 3 different prompts and their settings as detailed in \ref{A.1}. We also compared our results with \cite{Han2023IsIE}, who investigated the performance of ChatGPT on various information extraction tasks, including ABSA. 
All results are shown in table \ref{table44}. From the experiment, we find that when we prompt ChatGPT with some instructions (single aspect and multi-aspect sentences) can better improve the performance, but its best results are still not as good as our model. This situation suggests that ChatGPT possesses the ability to understand fine-grained segmental information to some extent. Perhaps there are ways to better harness this ability, such as incorporating constituent trees and dynamic local attention mechanisms as described in this paper (refer \ref{A.1} for details).

\subsection{Impact of Constituent Tree Layer Number}
\begin{figure}[H] 
\centering 
\begin{minipage}[t]{0.3\columnwidth}
\centering
\begin{tikzpicture}[scale=.40]
\begin{axis}[
    xlabel={\textbf{Layer number of constituent tree}},
    ylabel={\textbf{Accuracy.($\%$)}},
    xmin=1.8, xmax=5.2,
    ymin=74, ymax=92,
    xtick={2,3,4,5},
    ytick={74,76,78,80,82,84,86,88,90},
    legend pos=north west,
    ymajorgrids=true,
    grid style=dashed,
]

\addplot[
    color=red,
    mark=diamond*,
    ]
    coordinates {
    (2,76.95)
    (3,76.8)
    (4,77.84)
    (5,74.59)
    };
\addplot[
    color=blue,
    mark=triangle*,
    ]
    coordinates {
    (2,85.61)
    (3,86.23)
    (4,87.31)
    (5,85.07)
    };
\addplot[
    color=green,
    mark=*,
    ]
    coordinates {
    (2,79.3)
    (3,80.72)
    (4,82.46)
    (5,79.77)
    };
    \legend{Twitter, Restaurant, Laptop}
    \end{axis}
\end{tikzpicture}
\end{minipage}
\quad
\begin{minipage}[t]{0.63\columnwidth}
\centering
\begin{tikzpicture}[scale=.40]
\begin{axis}[
    xlabel={\textbf{Layer number of constituent tree}},
    ylabel={\textbf{F1.($\%$)}},
    xmin=1.8, xmax=5.2,
    ymin=73, ymax=86,
    xtick={2,3,4,5},
    ytick={74,76,78,80,82,84},
    legend pos=north west,
    ymajorgrids=true,
    grid style=dashed,
]
\addplot[
    color=red,
    mark=diamond*,
    ]
    coordinates {
    (2,76.02)
    (3,75.35)
    (4,77.11)
    (5,73.52)
    };
\addplot[
    color=blue,
    mark=triangle*,
    ]
    coordinates {
    (2,79.13)
    (3,80.05)
    (4,82.84)
    (5,78.48)
    };
\addplot[
    color=green,
    mark=*,
    ]
    coordinates {
    (2,76.14)
    (3,77.11)
    (4,79.07)
    (5,76.53)
    };
    \legend{Twitter, Restaurant, Laptop}
    
\end{axis}
\end{tikzpicture}
\end{minipage}
\caption{Effect of the number of constituent tree layers.} 
\label{layer number of constituent tree} 
\end{figure}
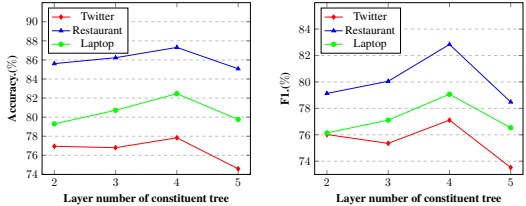

To investigate the impact of different layer numbers of the constituent tree, we evaluate the performance of the model with 2 to 5 constituent tree layers on three different datasets. 
As shown in Figure \ref{layer number of constituent tree}, the best performance of the model is achieved when the number of layers of the constituent tree is 4.
When the layer numbers of the constituent tree are lower than 4, the information from the constituent tree cannot fully cover the entire sentence, resulting in the model not being able to fully learn the complete segment information of a sentence. When the layer numbers are greater than 4, the model will 
repetitively learn redundant segment information, resulting in a decrease in model performance.

\subsection{Case Study}
We conduct a case study with a few examples, shown in Table \ref{table4}.
The first sentence only has one 
aspect word, so all models can easily determine the sentiment polarity of the aspect correctly. For the second comparative type of sentence, there is a certain syntactic dependency between the aspect "hamburger with special sauce" and the aspect "big mac", and the SeSG, which lacks syntactic information, cannot handle this type of sentence well.
The last cases contain multiple aspects and opinions. DualGCN and SyLG, which lack segment structure awareness, cannot focus on local information around aspects, resulting in incorrect judgments.
Our S$^2$GSL correctly predicts all samples, indicating that it effectively considers the complementarity between segment structures and syntax correlations of a sentence.

\section{Related Work}

With the rapid development of ABSA, current research can be broadly divided into three main categories attention-based methods, syntactic-based methods, and multi-graph combined methods.

\textbf{Attention-based methods}
Recently, various attention mechanisms have been proposed to implicitly construct the semantic relationships between aspects and their context. \cite{wang-etal-2016-attention,tang-etal-2016-aspect,2017arXiv170900893M,chen-etal-2017-recurrent,gu-etal-2018-position,fan-etal-2018-multi,hu-etal-2019-constrained,tan-etal-2019-recognizing}.
For instance, \cite{wang-etal-2016-attention} proposed an attention-based Long Short-Term Memory (LSTM) network for aspect-based sentiment classification.
\cite{2017arXiv170900893M} proposed an interactive attention network that can model the connection between the target aspect and the context simultaneously.
\cite{hu-etal-2019-constrained} propose orthogonal regularization and sparse regularization so that the attention weights of multiple aspects can focus 
\noindent on different parts of a sentence.

\textbf{Syntactic-based methods}
Several studies have explicitly used syntactic knowledge to explicitly build connections between aspects and opinions.\cite{dong-etal-2014-adaptive,zhang-etal-2019-aspect,sun-etal-2019-aspect,huang-carley-2019-syntax,zheng2020replicate,wang-etal-2020-relational,zhou-etal-2021-closer}.
\cite{wang-etal-2020-relational} reshape the conventional dependency tree with manual rules so that the root node of the dependency tree points to the aspect word.
\cite{zhou-etal-2021-closer} constructed a task-oriented latent tree in an end-to-end fashion.

\textbf{Multi-graph combined methods}
With the rapid growth of Graph Convolutional Neural Networks, some studies have explored the combination of different types of graphs for ABSA.
For example, \cite{chen-etal-2020-inducing} used a gating mechanism to combine a dependency graph and a latent graph to generate task-oriented representations.
\cite{li-etal-2021-dual-graph} constructed two graph convolutional neural networks using dependency tree and attention mechanism.
\cite{tang-etal-2022-affective} by constructing a latent graph and a semantic graph to effectively capture the interaction between aspects and distant opinions.
\cite{liang-etal-2022-bisyn} proposes to simultaneously utilize constituent tree and dependency tree of a sentence to model the sentiment relations between each aspect and its context.

However, these methods have primarily relied on the global graph structure learning process, which tends to introduce irrelevant contextual information and syntactic dependency unrelated to specific aspects.
Our proposed method of using two graph branches can effectively align each aspect word with its corresponding opinion word.
 
\begin{table*}[htb]
\centering
\scalebox{0.80}{\begin{tabular}{lcccc} 
\toprule
\textbf{Sentences}                                                                                                                                                   & \textbf{DualGCN} & \textbf{SyLG} & \textbf{SeSG} & \textbf{S$^2$GSL}  \\ 
\hline
1. Much more reasonably $\left[\textbf{priced}\right]_{p}$ too!                                                                                                                      & (P$\checkmark$)              & (P$\checkmark$)            & (P$\checkmark$)            & (P$\checkmark$)             \\ 
\hline
\begin{tabular}[c]{@{}l@{}}2. New $\left[\textbf{hambuger with special sauce}\right]_{p}$ is ok - at least better \\than $\left[\textbf{big mac}\right]_{n}$!~\end{tabular}                          & (P$\checkmark$,O$\times$)          & (P$\checkmark$,N$\checkmark$)         & (P$\checkmark$,P$\times$)          & (P$\checkmark$,N$\checkmark$)          \\ 
\hline
\begin{tabular}[c]{@{}l@{}}3. Perfectly al dente $\left[\textbf{pasta}\right]_{p}$, not drowned in $\left[\textbf{sauce}\right]_{o}$ - - \\generous $\left[\textbf{portions}\right]_{p}$.\end{tabular}               & (P$\checkmark$,N$\times$,P$\checkmark$)       & (P$\checkmark$,P$\times$,P$\checkmark$)      & (P$\checkmark$,O$\checkmark$,P$\checkmark$)      & (P$\checkmark$,O$\checkmark$,P$\checkmark$)       \\
\bottomrule
\end{tabular}}
\caption{Case study experimental results of four different models.The aspect words are included in [], and p, n, and o represent the true "positive", "negative", and "neutral" sentiment polarities.}
\label{table4}
\end{table*}

\section{Conclusion}
In this paper, we propose an S$^2$GSL model to tackle global structures that will introduce irrelevant contexts and syntactic dependencies during the process of graph structure learning. We propose a SeSG branch that decomposes the ABSA complex graph structure learning problem into multiple local sub-structure 
learning processes by utilizing constituent trees.
Moreover, we propose a SyLG branch, a more learnable method to introduce syntactic dependencies into latent tree construction.
Finally, we devise a Self-adaptive Aggregation Network to realize the interaction between two graph branches, achieving complementarity across diverse structures. Experiments on four benchmarks demonstrate S$^2$GSL outperforms the baselines. 

\section*{Limitations}
S$^2$GSL framework constructs different branches for syntactic and semantic structures which cannot encompass diverse structures in a unified graph modeling process. Therefore, the S$^2$GSL framework further devises an adaptive aggregation to fuse diverse structural information.


\section*{Acknowledgements}
This research was supported in part by the National Science and Technology Major Project (2021ZD0111501), the National Science Fund for Excellent Young Scholars (62122022), the Natural Science Foundation of China (62206064), the Guangzhou Basic and Applied Basic Research Foundation (2024A04J4384), the major key project of PCL (PCL2021A12), the Guangdong Basic and Applied Basic Research Foundation (2023B1515120020), and the Jihua laboratory scientific project (X210101UZ210). 

\bibliography{ref}

\renewcommand\thesection{\Alph{section}}
\setcounter{section}{0}
\maketitle

\section{Appendix}
\subsection{Experiments With ChatGPT}
\label{A.1}
The rise of large language models (LLMs) such as GPT-3 \cite{NEURIPS2020_1457c0d6}, PaLM \cite{Chowdhery2022PaLMSL}, Llama \cite{Touvron2023Llama2O}, etc, has greatly facilitated the rapid development of natural language processing (NLP). ChatGPT \cite{openai2023}, powered by GPT-3.5 and GPT-4, can achieve significant zero-shot and few-shot in-context learning (ICL) \cite{Brown2020LanguageMA} performance on unseen tasks, even without any parameter updates.

In this section, we conducted exhaustive experiments to investigate the performance of ChatGPT on ABSA tasks and its ability for fine-grained understanding of segmental context.

\begin{figure*}[!htb] 
\centering 
\includegraphics[width=2.0\columnwidth]{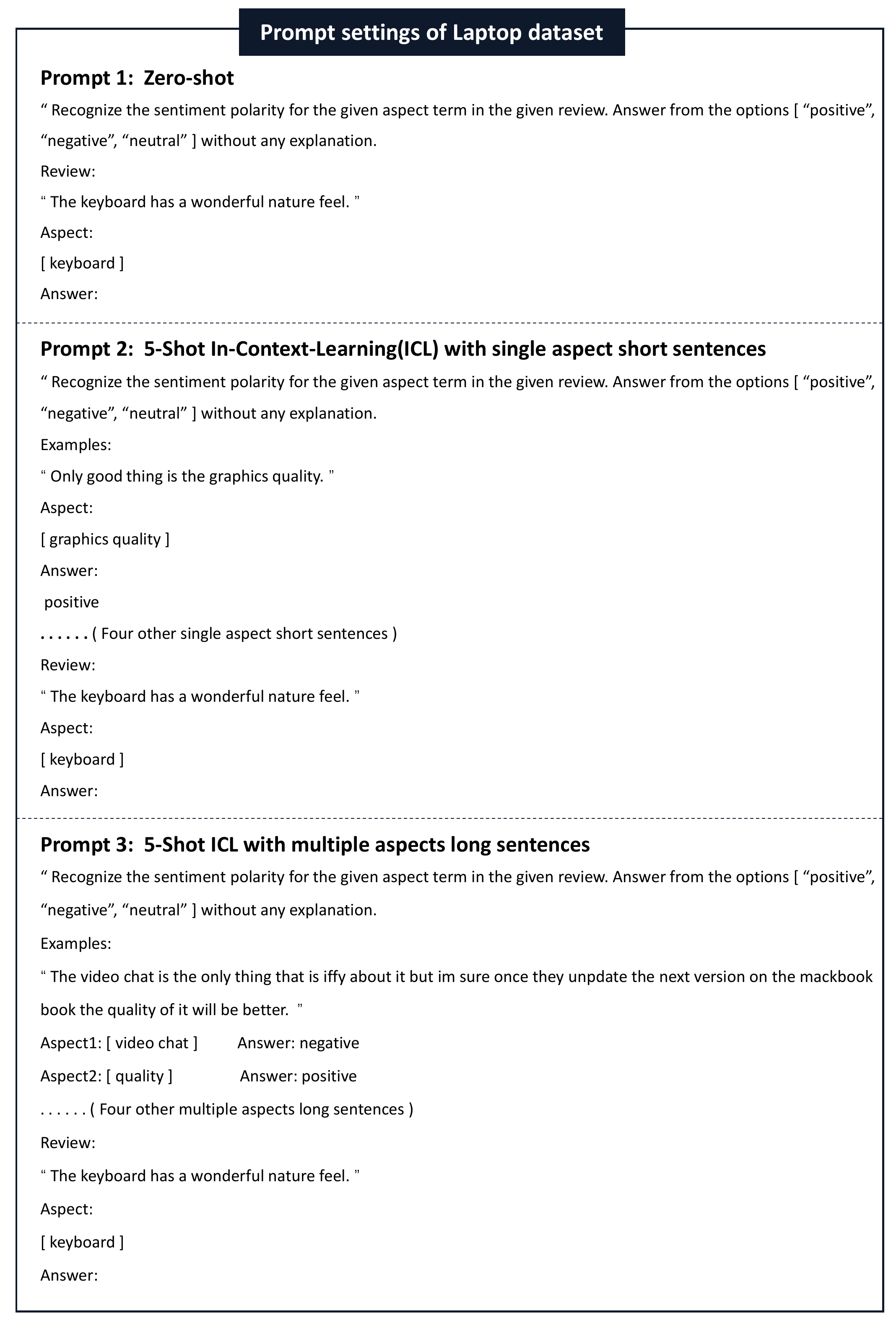} 
\caption{Prompt settings of laptop dataset.} 
\label{prompt-setting} 
\end{figure*}

\textbf{Experiment setting :} The version of ChatGPT we utilized in the experiment is \emph{gpt-3.5-turbo}. To avoid variations in ChatGPT-generated outputs, the \emph{temperature} parameter was set to 0. For the number of response words, the \emph{max\underline{~}tokens} parameter was set to 512. We experimented with 3 different prompts and their respective settings as shown in Figure\ref{prompt-setting}. We take the laptop dataset as example to explain our prompt settings:
\begin{itemize}
\item \textbf{Prompt-setting 1: Zero-shot.}  In this prompt setting, we have only given definition: "Recognize the sentiment polarity for the given aspect term in the given review. Answer from the options [ “positive”,
“negative”, “neutral” ] without any explanation.", then we sent a test review and all aspect words in this review to ChatGPT, and ChatGPT will give the answer. 

\item \textbf{Prompt-setting 2: 5-Shot ICL with single aspect short sentences.} The given definition is same as setting 1. What differs from setting 1 is that we randomly selected 5 single-aspect word sentences from the training set as examples to better leverage ChatGPT's In-Context Learning capability. Sequentially, we provide a test review and all aspect words, asking ChatGPT to output the sentiment polarity for each aspect word.

\item \textbf{Prompt-setting 3: 5-Shot ICL with multiple aspects long sentences.}
What differs from setting 2 is that the examples chosen from the training set are all multiple aspects long sentences. The purpose of doing this is to explore 
whether ChatGPT possesses the ability to understand fine-grained segmental context and the strength of this ability.

\end{itemize}

\textbf{Result analysis :}
The results using ChatGPT and our model are shown in table\ref{table1}. Firstly, from the result of prompt-setting 2, we can infer that using a few examples to guide ChatGPT can effectively improve its performance, demonstrating the powerful In-Context-Learning capability of ChatGPT. Secondly, we compared our experimental results with \cite{Han2023IsIE}, who investigated the performance of ChatGPT on various information extraction tasks, including ABSA. We found that our model has better performance. Finally, from prompt-setting 3, we can observe that providing ChatGPT with long examples containing multiple aspect words can lead to better performance. However, the results still fall short compared to our model. This situation suggests that ChatGPT possesses the ability to understand fine-grained segmental information to some extent. Perhaps there are ways to better harness this ability, such as incorporating constituent trees and dynamic local attention mechanisms as described in this paper.

\subsection{Paramer Comparison}
\label{A.4}
We conduct a parameter comparison between S$^2$GSL and other GCN-based baseline methods, shown in table \ref{table5}. Notably, the number of parameters in S$^2$GSL is comparable while the two graph learning branch design of S$^2$GSL does lead to an increase in the parameters, it's important to note that this does not result in an unnecessary escalation in computational costs when compared with the baselines.

\begin{table}[H]
\centering
\scalebox{0.8}{\begin{tblr}{
  cells = {c},
  cell{1}{1} = {r=2}{},
  hline{1,8} = {-}{0.08em},
  hline{2} = {2-3}{},
  hline{3,7} = {-}{},
}
\textbf{Model}       & \textbf{Laptop14} & \textbf{Restaurant14} \\
                     & \textbf{F1}       & \textbf{F1}           \\
5-shot ICL \cite{Han2023IsIE}           & 76.76             & 81.85                 \\
Prompt-setting 1     & 75.62             & 79.48                 \\
Prompt-setting 2     & 77.02             & 81.38                 \\
Prompt-setting 3     & 77.91             & 82.14                 \\
\textbf{S$^2$GSL(Ours)} & \textbf{79.09}    & \textbf{82.84}        
\end{tblr}}
\caption{The results obtained using ChatGPT and S$^2$GSL. All three prompt settings are described in the text.}
\label{table1}
\end{table}

\begin{table}[H]
\centering
\scalebox{0.65}{\begin{tblr}{
  cells = {c},
  cell{1}{1} = {r=2}{},
  cell{1}{2} = {r=2}{},
  hline{1,3,7-8} = {-}{},
  hline{2} = {3-5}{},
}
\textbf{Model} & \textbf{Parameter Count} & \textbf{Laptop} & \textbf{Restaurant} & \textbf{Twitter} \\
               &                          & \textbf{F1}     & \textbf{F1}         & \textbf{F1}      \\
ACLT(2021)     & 110M                    & 75.83           & 78.44               & 74.51            \\
T-GCN(2021)    & 113M                    & 77.03           & 79.95               & 75.25            \\
DualGCN(2021)  & 111M                    & 78.10           & 81.16               & 76.02            \\
SSEGCN(2022)   & 110M                    & 77.96           & 81.09               & 76.02            \\
\textbf{S$^2$GSL} & \textbf{114M}           & \textbf{79.07}  & \textbf{82.84}      & \textbf{77.11}   
\end{tblr}}
\caption{Paramer comparison with other GCN-based baseline methods. }
\label{table5}
\end{table}

\end{document}